# Leveraging Convolutional Sparse Autoencoders for Robust Movement Classification from Low-Density sEMG

**Blagoj Hristov\*, Zoran Hadzi-Velkov, Katerina Hadzi-Velkova Saneva, Gorjan Nadzinski, Vesna Ojleska Latkoska**

\*All authors are with the University "Ss. Cyril and Methodius", Faculty of Electrical Engineering and Information Technologies, Skopje, North Macedonia. Corresponding author: Blagoj Hristov (e-mail: hristovb@feit.ukim.edu.mk).

## ABSTRACT

Reliable control of myoelectric prostheses is often hindered by high inter-subject variability and the clinical impracticality of high-density sensor arrays. This study proposes a deep learning framework for accurate gesture recognition using only two surface electromyography (sEMG) channels. The method employs a Convolutional Sparse Autoencoder (CSAE) to extract temporal feature representations directly from raw signals, eliminating the need for heuristic feature engineering. On a 6-class gesture set, our model achieved a multi-subject F1-score of 94.3% ± 0.3%. To address subject-specific differences, we present a few-shot transfer learning protocol that improved performance on unseen subjects from a baseline of 35.1% ± 3.1% to 92.3% ± 0.9% with minimal calibration data. Furthermore, the system supports functional extensibility through an incremental learning strategy, allowing for expansion to a 10-class set with a 90.0% ± 0.2% F1-score without full model retraining. By combining high precision with minimal computational and sensor overhead, this framework provides a scalable and efficient approach for the next generation of affordable and adaptive prosthetic systems.

## Introduction

Achieving reliable functionality and ease of use is critical for the long-term adoption of artificial prosthesis among users. The device is far more likely to be used continuously if it offers intuitive and responsive control, while systems that are overly complex or ineffective are often quickly abandoned[1]. In order to deliver this type of responsive control, many modern prostheses leverage surface electromyography (sEMG) to capture muscle activation signals from the residual limb. These signals must then be accurately and quickly interpreted to determine the user's intended movement. The effectiveness of this classification procedure directly influences both the fluidity of the control and the range of possible motions that the device can support.

Another essential component to take into consideration in prosthesis design is the cost of the device. Many advanced prosthetic systems, while offering decent functionality and comfort, are extremely expensive and often priced in the thousands of US dollars[2], primarily due to the high number of EMG sensors, the type of materials used to build the device as well as the complexity of their processing systems.

The conventional framework for myoelectric control has historically relied on classification based on handcrafted features derived from time-domain characteristics (e.g., mean, variance) and frequency-domain analyses (e.g., power spectral density and frequency bands)[3]. This established paradigm was founded on the principle of leveraging expert domain knowledge to manually derive a low-dimensional feature vector from the complex, high-dimensional surface electromyography (sEMG) signal. While this handcrafted approach laid the essential groundwork for modern prosthetics and has proven effective for basic control tasks[4], it is constrained by its fundamental reliance on predefined, manually selected metrics. The process of defining an optimal feature set is often heuristic and labor-intensive, and there is no guarantee that these expert-defined features capture the complete, nuanced information present in the raw sEMG data. In fact, by reducing the signal to a handful of assumed characteristics, subtle yet potentially discriminative patterns may be inadvertently discarded[5]. This inherent limitation, the dependence on a fixed, potentially suboptimal feature engineering process, motivates a shift towards data-driven methods, like autoencoders, that can learn representations directly from the signal itself[6]. Such an approach promises to uncover more robust and expressive features, paving the way for more intuitive and performant control systems.

Standard Fully-Connected Autoencoders (FCAE) often fail when applied to raw sEMG time-series, yielding poor accuracy, and requiring complex preprocessing steps like transforming the signal into spectrogram or wavelet representations to achieve high performance[7]. We stipulate that a convolutional architecture, which inherently preserves and processes this temporal structure, is fundamentally better suited for this task and will significantly outperform architectures that discard this information by flattening the input signal. The success of Convolutional Neural Networks (CNNs) validates the use of convolutional architectures for learning directly from raw sEMG time-series[8,9]. However, the features learned by these supervised models are inherently task-specific and user-dependent, often requiring a full, data-intensive retraining process for every new user or set of gestures. This bias is a significant barrier to creating adaptable, real-world systems. We suggest that a more robust and flexible paradigm is to first create a generalized model of the sEMG signal itself. A convolutional autoencoder is uniquely suited for this role. By training in an unsupervised manner on diverse data from multiple subjects, the autoencoder learns to extract features that represent the fundamental physics of muscle contraction, rather than the unique patterns of a single individual. This process forms a powerful, generalized model of the neuromuscular activity[10].

Modern approaches have managed to preserve the temporal information of sEMG data by using Recurrent Neural Networks (RNNs), or their advanced variants like Long-Short Term Memory (LSTM). Their application for gesture recognition has been explored in[11], as well as their implementation in autoencoders for imputation of general multivariate time series[12]. However, autoencoders based on these types of networks carry significant disadvantages for real-time applications, including high computational complexity and susceptibility to the vanishing gradient problem[13]. The sequential processing inherent to RNNs, where each step depends on the previous one, creates a computational bottleneck that is ill-suited for the low-latency demands of prosthetic control or real-time human-computer interaction. While LSTMs were developed to mitigate the vanishing gradient issue, they do so at the cost of increased complexity and do not entirely solve the problem, potentially failing to learn long-range dependencies in the sEMG signal.

A Convolutional Autoencoder (CAE) is a more efficient alternative to RNNs and LSTMs, because it can capture local temporal patterns and hierarchical features in a highly parallelizable manner[14]. This approach avoids the complex recurrent connections of LSTMs, thereby reducing the computational load and making it more practical for implementation on resource-constrained hardware. A CAE thus achieves an optimal balance between performance and efficiency, providing a practical solution for real-world sEMG applications.

LASSO regularization is a powerful tool for encouraging sparsity in the output of all machine, including DNN, models[15]. By applying a L1 penalty to the bottleneck layer of a CAE, we expect the model to learn a more disentangled and robust set of features, leading to improved generalization and classification performance compared to an identical, architecture without LASSO regression. Additionally, sparsity acts as a powerful form of regularization against overfitting by reducing the model's effective capacity for learning. Sparse autoencoders have been shown to extract abstract features that lead to superior classification accuracy when used with sEMG data[16]. However, an overly strong penalty can be detrimental, forcing the model to discard essential information and degrading performance. Therefore, we postulate that an optimally tuned L1 penalty provides a beneficial regularizing bias, guiding the network to learn a refined feature set that outperforms a standard CAE.

A primary obstacle in myoelectric control is the high inter-subject variability of sEMG signals, which causes models trained on a group of subjects to generalize poorly to new users[17]. Transfer learning, and specifically fine-tuning, offers a powerful solution to this domain adaptation challenge[18]. While advanced strategies exist, such as replaying old data[19] or using generative models to synthesize it[20], our proposed framework aims for a more pragmatic approach requiring minimal data and computational processing power. We propose that by leveraging a pre-trained, generalized model as a robust feature extractor, we can fine-tune the parameters of a simple classifier to rapidly adapt to a new user's unique physiological signals through targeted, minimal adjustments to only the final decision-making layers of the network, using just a few new measurements. Furthermore, we seek to demonstrate that this same principle of efficient adaptation can enable a seamless expansion of the model's capabilities. Instead of undertaking a costly and time-consuming retraining of the entire network, our approach will be able to incrementally increase the number of recognized movements by strategically modifying the model's output stage. We anticipate that this method will preserve the core knowledge of the originally learned gestures, thus providing a flexible and scalable solution.

While autoencoders have already been applied to sEMG signals, existing methods present trade-offs that limit their practicality for the kind of lightweight, adaptable system we propose. For instance, while transfer learning has shown promise, challenges related to domain shift between subjects require robust adaptation strategies to maintain high performance[21]. One such approach can be seen in[22] where the authors propose a convolutional autoencoder for EMG signal compression, attaining classification accuracy of ~85.5% using eight EMG channels with a high compression ratio. To achieve this, their method relies on a large number of sensors and significant computational resources for training. Another similar use of autoencoders can be seen in[7] where the authors propose the use of a deep learning framework using stacked sparse autoencoders to extract features from sEMG signals represented via spectrograms, wavelets, and wavelet packets. Their system demonstrated high classification performance (up to 92.25% accuracy) across ten finger movements, while using two sensors and various different classifiers. However, the method emphasizes complex signal preprocessing and multiple feature representations, which only increases computational demand and execution time. Finally, another recent approach[23] employs a Variational Autoencoder (VAE) to disentangle input data into three distinct representations: subject-specific, gesture-specific, and residual features. While their system demonstrates superior generalization with an average accuracy of 90.52% in cross-subject validation, it relies on a complex model architecture that may not be as effective at capturing the inherent spatio-temporal structures of sEMG signals and operates on dense data from an eight-channel setup.

As a potential solution to these problems, we introduce a Convolutional Sparse Autoencoder (CSAE) architecture, trained in an unsupervised manner, for the extraction of features directly from raw, two-channel sEMG signals, combined with a simple neural network classifier for the classification of the movements. To rigorously evaluate the efficacy of the proposed approach, the methodology is comprised of a multi-stage validation process across two distinct datasets. Initially, the classifier's multi-subject performance is assessed on the first dataset, comprised of six distinct classes (five individual finger flexions and one hand-close gesture), by testing on new movement instances from subjects whose data is already included in the training set. Then, in order to evaluate its generalization capability, the model is fine-tuned using minimal data from a new, previously unseen subject, followed by evaluation on new movement instances from that same subject. Finally, the classifier is adapted to extend its recognition capabilities to the second, expanded 10-class dataset, encompassing both individual and combined finger flexions, and its performance is calculated on this broader set of movements. Based on this, the present work aims to provide the following contributions:

(1) Development of a lightweight, end-to-end deep learning framework that enables high-fidelity movement classification using only two raw sEMG channels. This approach challenges the prevailing reliance on overparameterized models and cumbersome high-density sensor arrays. By maintaining state-of-the-art accuracy with significantly reduced computational and hardware overhead, the framework offers a scalable, clinically viable alternative to current high-cost systems.
(2) Quantitative validation of a novel Convolutional Sparse Autoencoder (CSAE) architecture. Our analysis demonstrates that: (a) preserving the signal's temporal structure via 1D convolutions is critical for learning disentangled representations, and (b) an optimally tuned L1 sparsity constraint provides a superior regularizing bias, significantly improving feature robustness and cross-subject generalization compared to traditional dense or non-sparse models.
(3) Implementation of a robust adaptive framework that addresses two distinct clinical needs: (a) rapid personalization for new users through a low-latency, few-shot calibration protocol, and (b) continuous system evolution using an incremental learning strategy that seamlessly integrates new gestures while robustly mitigating catastrophic forgetting.

## Methods

**Data preprocessing**

To train classifiers capable of distinguishing between the distinct movements, it is essential to first collect the relevant data for training. This is accomplished by using two sEMG sensors to monitor the electrical activity in the user's forearm. The datasets employed in this study, originally described by Khushaba et al.[24], include data collected and sampled at 4kHz from eight volunteers, where the data acquisition was conducted in accordance with relevant guidelines and regulations. As reported by the original authors, informed consent was obtained from all participants, and the study was conducted under the ethical guidelines of the University of Technology, Sydney. This retrospective study utilized only previously collected and fully anonymized sEMG data. In accordance with the Law on Health Protection and the Law on Personal Data Protection of the Republic of North Macedonia, ethical approval is typically required for research involving identifiable human subjects or personal health data. However, since no identifiable data were used and all analyses were conducted on publicly available, anonymized datasets, ethical approval by an institutional review board was deemed unnecessary.

Two Delsys DE 2.x sensors were placed on the flexor digitorum profundus muscle, which controls finger flexion, and the extensor digitorum muscle, responsible for finger extension. For each type of movement, six separate 5-second-long trials were recorded. Each trial consists of a sustained isometric contraction held at a moderate force level. To prevent muscle fatigue, a rest interval of approximately 3-5 seconds was provided between consecutive repetitions. The data was recorded in a single continuous session per subject to minimize electrode shift.

Two separate datasets are used in this paper. Dataset 1 consists of six distinct movements: five individual finger flexions and a single hand-close grip motion (formation of a fist). Dataset 2 contains the same types of movements as in the first one plus an additional four combined flexions (thumb-to-finger touches), for a total of ten unique classes. The performed gestures are presented in Figure 1. The use of two distinct datasets allows for a direct evaluation of the model's performance against increasing task complexity and its ability to adapt to new, unseen classes.

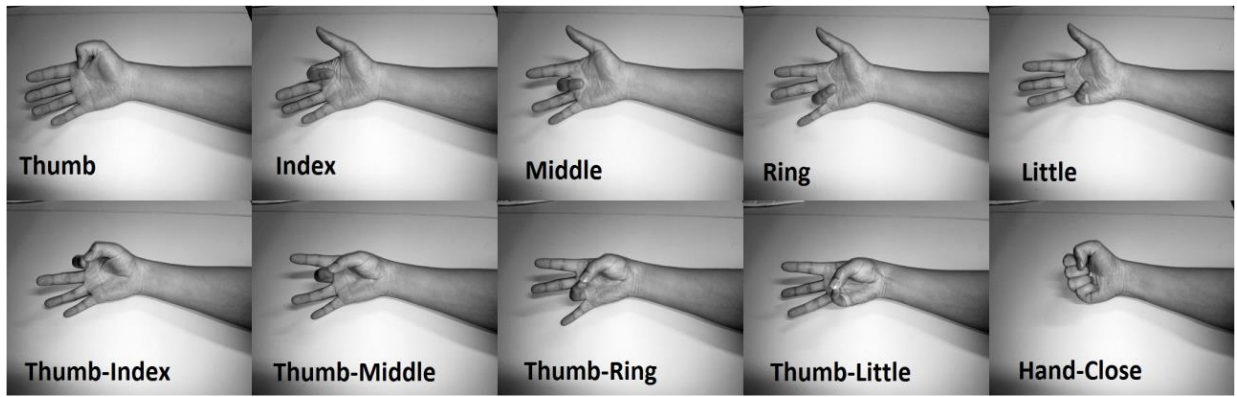

**Figure 1**. Illustration of the dynamic gesture protocols. The top row displays individual finger flexions, while the bottom row depicts thumb-opposition pinches. Dataset 1 is composed of the five individual flexions and the gross hand-close gesture. Dataset 2 encompasses the complete ten-class set, combining the elements of Dataset 1 with the four thumb-opposition movements.

*Signal segmentation*

To be able to design a functional prosthetic device real-time control is essential, with only a minimal acceptable delay in carrying out the desired movement. In order to achieve this, each measurement is segmented into shorter fragments of 250ms, as this is the maximum delay that can be tolerated without being noticeable by the user,[25]. Maximizing the amount of data stored in each segment is done through the use of an overlapping sliding window approach, where a 250ms length window with a 125ms stride (50% overlap) slides across the 5-second signals to create short segments of the data. This segmentation strategy yields several benefits. The 50% overlap doubles the effective number of training samples and, critically, preserves the temporal dependency between adjacent segments, which can help improve classification accuracy. Operationally, this allows for a new classification decision to be made every 125ms while still analyzing a full 250ms of recent data. The resulting 125ms interval between classifications provides a practical processing window for the prosthesis's actuators to activate and position the fingers. Therefore, the main objective of the classifier is to distinguish between these individual 250ms segments, each corresponding to a certain type of movement, rather than classifying the entire 5-second measurement of that same movement, which would be impractical for real-world use.

*Dataset Partitioning*

To ensure rigorous evaluation and prevent potential data leakage, the dataset was partitioned into distinct sets using a leave-one-subject-out (LOSO) cross-validation protocol. As illustrated in Figure 2, this protocol is divided into two distinct phases separated by a strict data isolation barrier:
1) **Multi-Subject Classification (source domain):** In each fold, data from seven subjects constitutes the source domain. To evaluate the general model's baseline performance on seen users (closed-set evaluation), the data from these seven subjects is partitioned chronologically: trials 1 – 4 are allocated for training, trial 5 is used for hyperparameter tuning (validation), and the final trial 6 is held out for performance evaluation.
2) **User-Specific Adaptation (target domain):** The eighth subject is held out entirely as the unseen target. To evaluate the efficiency of the transfer learning protocol, this subject's data is split into a minimal calibration set from trial 1, a validation set for hyperparameter tuning from trial 2, and a final held-out test set from trials 3 – 6 to assess the adapted model's performance.

This process was repeated eight times, with each subject serving as the target domain exactly once. The results represent the average performance and standard errors across all eight folds for both the general and adapted models.

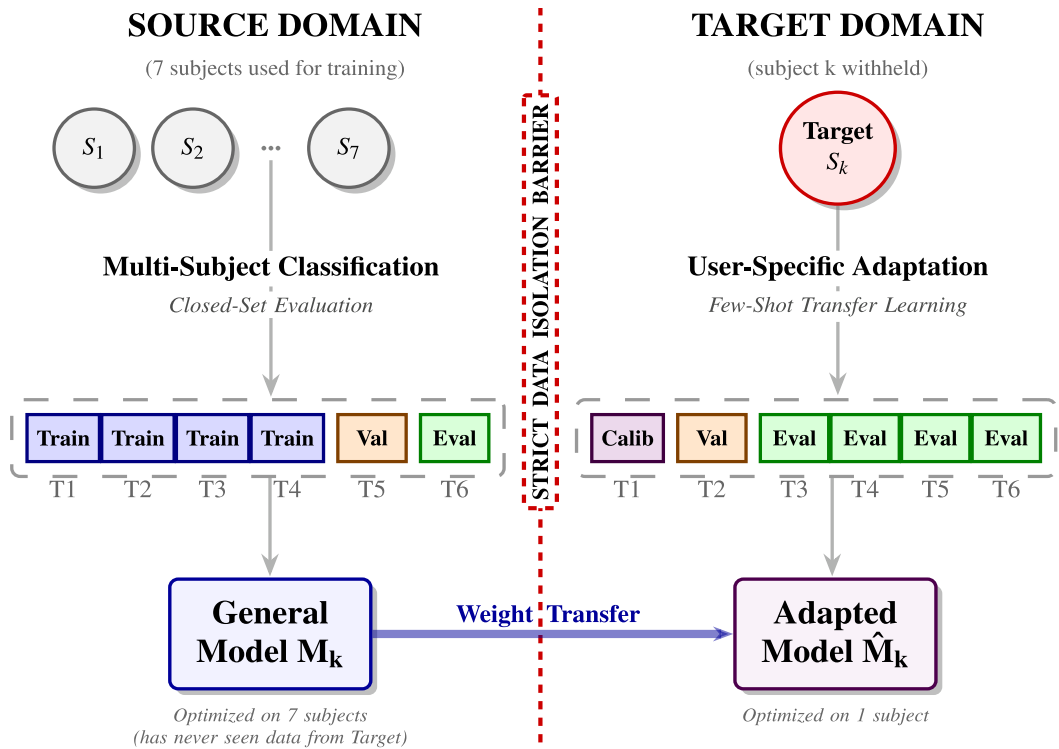

**Figure 2.** Schematic overview of the proposed data partitioning framework. The framework employs a leave-one-subject-out strategy divided into two phases. Each subject is represented by 6 trials T1-T6. (Left) Multi-Subject Classification (source domain): A general model is trained on a pooled dataset from seven subjects (S1−S7). Trials T1−T4 are utilized for weight optimization, T5 for validation, and T6 for closed-set evaluation, establishing the baseline capability on seen users. (Right) User-Specific Adaptation (target domain)**:** The pre-trained weights are transferred to a completely unseen target subject. The model undergoes few-shot calibration using trial T1, validation on T2, and final evaluation on the remaining unseen trials T3−T6.

*Standardization*

Prior to model training, the segmented data underwent a crucial standardization step. To ensure no information leakage from the validation or test sets, the standardization scaler was fitted only on the training dataset. The mean and variance learned from this training data were then used to transform both the validation and test datasets, ensuring that the model was evaluated under realistic conditions that mimic a deployed prosthesis control system processing new, unseen data. As the same initial classifier model is also used in the fine-tuning and new-class adaptability evaluations on the two other datasets, the learned parameters were used to transform those datasets as well.

**Feature extraction with a convolutional sparse autoencoder**

The accurate and efficient interpretation of sEMG signals is central to achieving real-time multi-movement prosthesis control. Traditional EMG classification pipelines usually rely on statistical time-domain and frequency features, such as the widely adopted Hudgins feature set[3], which are often limited by the domain expertise of the researcher and may not capture the full complexity of the underlying muscle activation patterns. To overcome these limitations, we employ a data-driven approach for representation learning.

We propose a Convolutional Sparse Autoencoder (CSAE), a deep neural network trained in an unsupervised manner to learn a compact and meaningful representation of the sEMG signal. Unlike classical methods like Convolutional Sparse Coding (CSC), which require complex and computationally expensive optimization procedures, the autoencoder framework allows for efficient end-to-end training using standard backpropagation. The model, illustrated in Figure 3, is trained exclusively on the training data partition from the seven subjects to learn a general feature representation, which is later used by a downstream classifier.

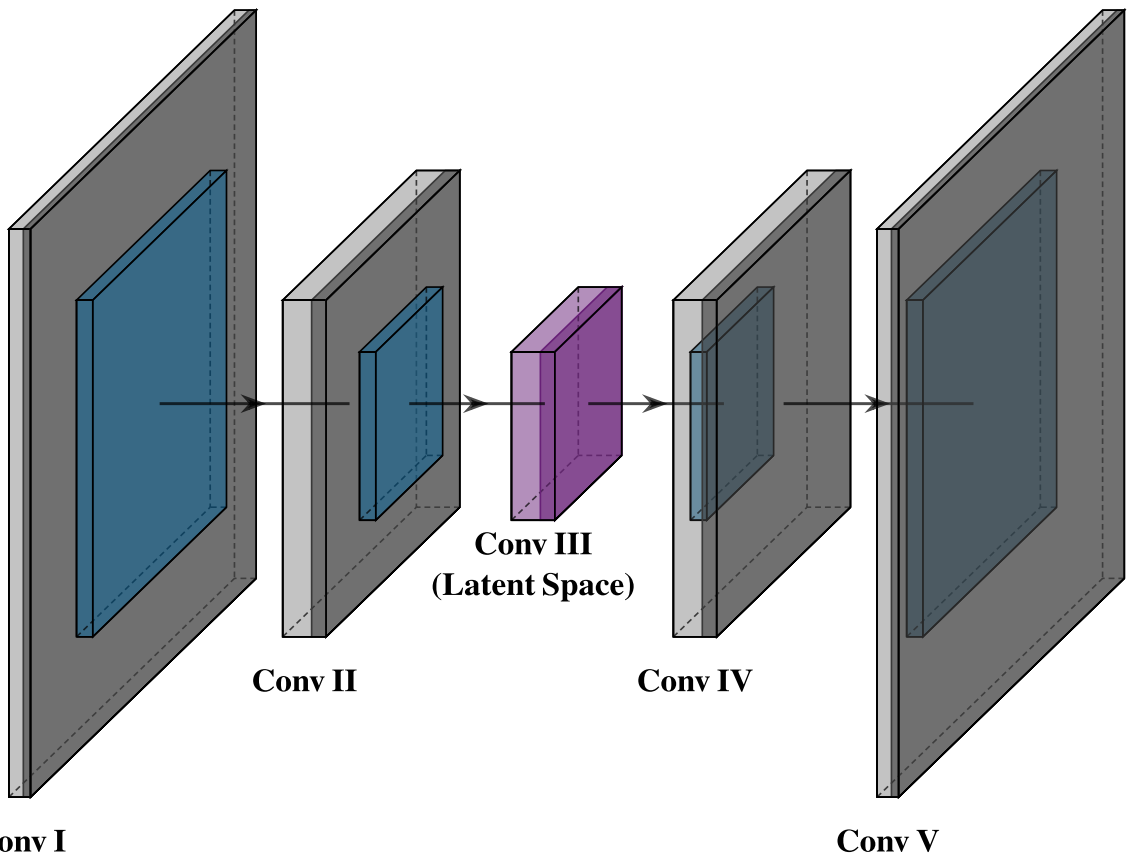

**Figure 3.** Architecture of the proposed Convolutional Sparse Autoencoder. The encoder (left) progressively downsamples the input signal through two one-dimensional convolutional blocks (Conv I, Conv II) into a compressed feature representation at the bottleneck layer (Conv III, purple). The decoder (right) then symmetrically upsamples this representation through the corresponding blocks (Conv IV, Conv V) to reconstruct the original signal from the latent features.

*CSAE Architecture*

Let the standardized sEMG input signal be represented by a matrix $X \in \mathbb{R}^{T \times C}$, where $T = 1000$ denotes the number of time samples per segment (250ms at 4 kHz), while $C = 2$ is the number of channels (sensors). The objective is to train the autoencoder, composed of an encoder function $f_\phi$ and a decoder function $g_\theta$, to learn a latent representation $Z = f_\phi(X)$, where $Z \in \mathbb{R}^{T' \times D} (T' < T, D$ is the number of latent filters).

This learned representation should ideally satisfy three key properties:

1) **Information Preservation:** $Z$ must retain the essential characteristics of the signal required to distinguish different movements.
2) **Unsupervised Learnability:** The parameters $\phi$ and $\theta$ must be learnable solely from the input data $X$, without requiring any class labels.
3) **Sparsity:** The latent representation $Z$ should be sparse. This is a theoretically motivated choice designed to encourage the network to learn a disentangled and efficient representation of the input data where individual features are selective for specific patterns, which can improve model generalization[26]. By applying an L1 penalty to the activations in the bottleneck layer, we force the model to represent the signal using fewer strong activations, rather than a denser combination of many.

The proposed model is a fully convolutional autoencoder comprised of a symmetrical encoder-decoder pair designed to balance temporal structure retention, computational efficiency and sparsity.

*Encoder*

The encoder part of the autoencoder is where the main processing of the data occurs and it is used to compress the input signal $X$ into a low-dimensional latent sequence $Z$ while preserving its temporal structure. This is achieved through a hierarchy of convolutional blocks.

Each block $l$ in the encoder consists of three sequential operations: a one-dimensional convolution, downsampling, and an adequate activation function. As an alternative to fixed max-pooling operations, a key choice in our design is the use of strided ($S > 1$) convolutions for learnable downsampling, which is a modern approach used to reduce the temporal dimension $T$ at each stage.

While max-pooling is effective in most cases, it is a static, non-learnable operation with a rigid rule: to propagate only the maximum value within a receptive field. This can lead to a loss of information, as other features within the window are discarded regardless of their potential importance. In contrast, a strided convolution performs feature extraction and downsampling in a single, unified step. By using stride, the convolutional filter jumps across the input sequence, computing its output only at these discrete intervals. This process is inherently learnable such that through backpropagation, the network adjusts the filter weights not only to detect important patterns but also to ensure that the output of each stride is the most useful possible summary of the window it had just processed. The network itself learns the optimal downsampling function for the given task, rather than being constrained to a pre-defined maximum operation. This technique has been shown to be highly effective in modern all-convolutional network designs, often leading to better performance and a simpler architecture[27].

After each convolutional layer, a non-linear activation function is applied. Through empirical means, the best activation function for the task at hand was determined to be a leaky ReLU function, which can be expressed as:

$$LeakyReLU(x) = \begin{cases} x, & x > 0 \\ \alpha x, & x < 0 \end{cases}. \tag{1}$$

By implementing a small positive constant $\alpha$ we prevent the "dying ReLU" problem, as this allows for a small non-zero gradient when the neuron unit is not active, while also introducing non-linearity which enables the network to model complex data distributions.

The encoder architecture consists of two main downsampling blocks, which culminate in a final bottleneck layer designed to produce the compressed latent representation. This bottleneck is a one-dimensional convolutional layer that preserves the temporal structure of the feature maps, at a reduced resolution. It is within this layer that we introduce a sparsity constraint, which is a central component of our proposed method.

An L1 activity regularization term is applied to the activations of the bottleneck layer. This is a theoretically motivated choice, drawing from the principles of LASSO regression[15], designed to encourage the network to learn a disentangled and efficient representation of the input data. The regularization adds a penalty to the autoencoder's main loss function that is proportional to the sum of the absolute values (L1-norm) of the feature activations in the latent space. Mathematically, if **Z** is the output tensor of the bottleneck layer which contains the latent feature representation, the sparsity penalty $L_{reg}$ is defined as:

$$L_{reg} = \lambda \sum_i |Z_i|, \qquad (2)$$

where $Z_i$ are the individual activation values within the tensor **Z**, while $\lambda$ is the regularization coefficient that controls the strength of the penalty. The primary motivation for enforcing sparsity is to encourage the autoencoder to learn a more robust and interpretable set of features. We expect that there is an optimal value of $\lambda$ that encourages meaningful feature disentanglement without destroying essential information.

*Decoder*

The output of the bottleneck convolutional layer is the feature matrix $Z = f_\phi(X)$ that is used to classify the type of movement performed by the subject. In order to extract the correct features which carry important information about the data, first the autoencoder must be properly trained to accurately recreate the input, which is done by the decoder part of the encoder-decoder pair. The decoder is used to reconstruct the input signal $X$ from the compressed latent space **Z** by using the same architecture of the encoder, only mirrored in reverse. In order to revert the downsampling performed by the strided convolutions in the encoder, the decoder utilizes transposed convolutions with the same stride value as in the encoder. A transposed convolution is a learnable upsampling operation that effectively reverses the connectivity of a standard convolution, progressively increasing the temporal dimension of the feature maps while decreasing their channel depth. Similar to the encoder, a leaky ReLU activation is used, operating on the upsampled feature maps. The decoders learn to refine the upsampled representation in order to progressively reconstruct the input signal from the compressed features. The final convolutional layer in the decoder generates the reconstructed two channel EMG segment $\hat{X} = g_\theta(Z)$, which has the same dimensions as the original input $X$.

*Training Loss Function*

The autoencoder model is trained end-to-end by minimizing a Mean Squared Error (MSE) loss function that quantifies the difference (error) between the original input $X$ and the reconstruction $\hat{X}$. This is combined with the L1 activity regularization penalty on the bottleneck layer, yielding the final loss function:

$$L(\phi, \theta) = \frac{1}{N} \sum_{i=1}^{N} \left\| X_i - g_\theta\left(f_\phi(X_i)\right) \right\|^2 + \lambda \left\| f_\phi(X_i) \right\|, \qquad (3)$$

where N is the number of samples in the batch, and $\lambda$ is the regularization coefficient. The goal of the training procedure is to tune the weights of the layers such that the two signals are completely identical. The weights of the network are first initialized using a He-Normal distribution, and are then optimized using the AdamW algorithm. To ensure robust training and prevent overfitting, an early stopping procedure is also used where the training halts after validation performance no longer improves after a certain amount of training epochs, as well as a learning rate scheduler which gradually reduces the learning rate during training when validation loss stops improving for a certain number of epochs, which promote smooth convergence.

Upon completion of the training procedure, the model demonstrated high reconstruction fidelity, achieving an $R^2$ score of ~99%. This result, visually confirmed in Figure 4, is particularly significant given the constraints imposed on the network to learn a compressed representation. The degree of compression is determined by two key factors: the structural capacity of the bottleneck, defined by the number of latent filters, and the induced sparsity, controlled by the L1 regularization coefficient $\lambda$.

The choice of these parameters presents a critical trade-off between reconstruction fidelity and feature quality. A smaller bottleneck or higher sparsity forces the model to learn a more efficient, regularized representation, but risks discarding valuable information. Conversely, a larger bottleneck or lower sparsity can capture more detail but may lead to overfitting and less discriminative features. The optimal combination is not theoretically obvious and depends on finding a feature set that, while good enough for reconstruction, is ultimately most effective for the downstream classification task. Therefore, in order to determine the ideal parameters, a systematic hyperparameter search was conducted, evaluating the final classification performance for various combinations of latent filter sizes and L1 regularization strengths. The results of this empirical analysis are detailed in the Results section.

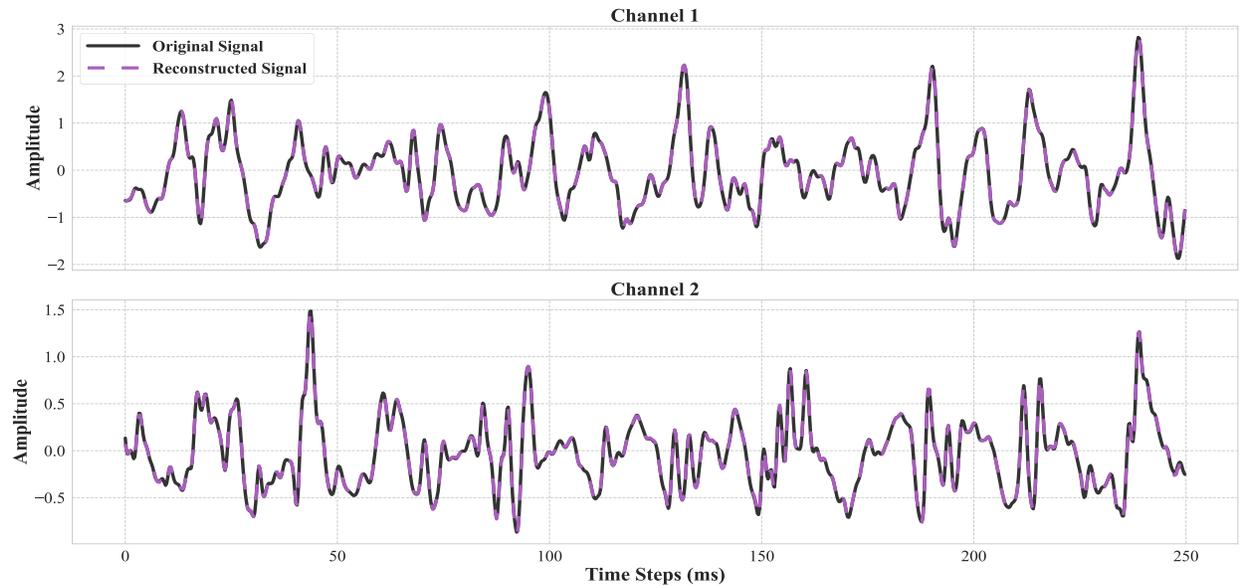

**Figure 4.** Example of sEMG signal reconstruction for a 250ms segment. The original signal (black, solid) and its corresponding reconstruction from the trained autoencoder (purple, dashed) are shown for both channels. The high degree of similarity and low reconstruction error visually confirm that the learned latent representation effectively captures the important characteristics of the input signal.

## Classification process

Following the unsupervised feature extraction phase with the CSAE, a dedicated classification model was trained to determine the type of movement the subject wants to perform from the learned EMG representations. This process employs a transfer learning approach, leveraging the rich feature matrix captured by the encoder.

*Classifier Architecture*

For the task at hand, a simple neural network was developed, the architecture of which is illustrated in Figure 5. First, the encoder part of the CSAE was extracted and its weights were frozen. The output of the encoder, a time-series feature map, is then passed through a specialized classifier head designed to interpret these representations. The classifier head first normalizes the features using a layer to ensure stable training. This step is critical for mitigating the effects of inter-trial and inter-subject signal variance.

The normalized features are then processed by a single one-dimensional convolutional layer to extract higher-level temporal patterns from the latent feature sequence. Instead of using a simple pooling operation which can lead to information loss, we implemented a self-attention mechanism to intelligently summarize the feature sequence. This mechanism calculates an importance score for each time step, creating a weighted context vector that emphasizes the most discriminative parts of the signal. The vector is then processed by a two-stage Multi-Layer Perceptron (MLP) head, which performs the final non-linear mapping required for classification. Finally, the entire model culminates in a softmax output layer which produces the probability distribution for the distinct movement classes.

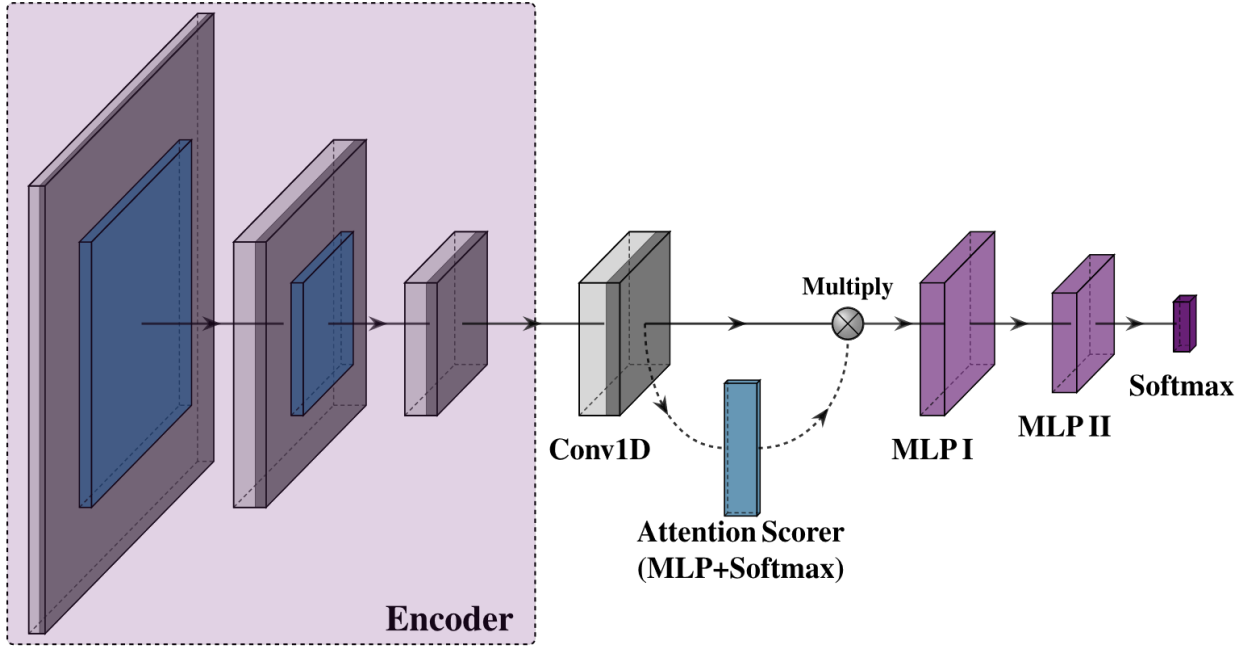

**Figure 5.** Architecture of the proposed classifier. The latent features from the frozen encoder are first processed by a single one-dimensional convolutional layer (Conv1D, gray) to further extract temporal patterns. A simple self-attention mechanism (blue) is used to calculate importance weights for each time step, which are then applied to the convolutional features to form a weighted context vector. The resulting vector is passed through a two-stage Multi-Layer Perceptron (MLP, purple), followed by a softmax layer (deep purple) which produces the final class probabilities.

Preliminary investigations into simpler, shallow convolutional neural networks (CNNs) without the autoencoder pre-training consistently failed to converge to F1-scores above ~90% on this low-density dataset, further justifying the use of the deeper, unsupervised feature extraction pipeline. To validate the use of the self-attention module, we conducted an ablation study comparing it against a standard Global Average Pooling (GAP) operation. The substitution of the attention mechanism with GAP resulted in an average performance degradation of ~1% in the F1-score, confirming that the adaptive weighting of temporal features yields a slight improvement in the model's ability to discern subtle hand movements, with comparable model complexity and inference time.

*Training and Evaluation*

Initially, the model was trained on the 6-class dataset using the pre-partitioned data from seven subjects (trials 1-4 for training, trial 5 for validation). The first objective was to establish a baseline performance on unseen data from users that were part of the original training pool. The loss function that was minimized during the training was the categorical cross-entropy loss, defined as:

$$L_{CE} = -\sum_{i=1}^{K}(y'_i)\log(\hat{p}_i), \qquad (4)$$

which calculates the difference between the predicted softmax probabilities $\hat{p}_i$ and the true one-hot encoded labels $y'_i$, for all six labels $K$. The training procedure was similar to the procedure used for training the autoencoder, employing AdamW as the optimization algorithm and early stopping and a learning rate scheduler to further prevent overfitting on the training data. The model's performance was then evaluated on the held-out test set (trial 6) from the same seven users.

*User-Specific Adaptation*

A critical requirement of any prosthesis control system is the ability to quickly adapt to a new user. To evaluate if the model is capable of this, the pre-trained classifier was tested on data from an eighth, completely unseen subject. This process involved a fine-tuning procedure, where trial 1 was used as a training set, trial 2 was used for validation, while the other 4 trials were used to evaluate the performance of the tuned classifier.

The fine-tuning phase focused on adapting only the highest-level decision-making layers of the network. First, all of the layers except for the final fully-connected layers were frozen, just like the encoder layers. By keeping the feature generation layers (the encoder and the classifier's convolutional layer) frozen, we ensure their robust, general knowledge is preserved. The only change in weights will occur in the fully-connected layers, which will be responsible for interpreting and combining the generated features such that they can accurately predict new unseen data from an entirely new user. This focused adaptation allows for the adjustment to the decision boundaries in the final output layer with minimal disruption to the established feature hierarchy in the previous layers of the network, while needing minimal data and computational cost, which is essential for a practical calibration scenario. Comparative experiments where the entire classifier head was unfrozen (while maintaining a frozen encoder) resulted in a ~1% average drop in performance. This confirms that the temporal features learned by the classifier's convolutional layer are robust across subjects, and fine-tuning them on the limited calibration data leads to overfitting. Therefore, the frozen strategy was deemed optimal, offering superior accuracy with significantly reduced training time. The model was then re-compiled with the saved weights from the initial training phase, and the tuning stage was initiated on the new data, using the same optimization procedure as in the initial training phase.

*Knowledge Transfer for New Movement Integration*

To evaluate the model's ability to expand its functionality without complete retraining, we adapted the fully trained 6-class classifier to a more complex 10-class dataset. This was achieved using a two-phase adaptation protocol designed to efficiently learn the new movements while mitigating the risk of catastrophic forgetting of the original knowledge.

First, the base classifier was loaded, its final 6-unit softmax output layer was removed, and a new randomly initialized 10-unit softmax output layer was appended to the decapitated model. To retain the knowledge of the original classes, the learned weights and biases for the six original movements were extracted from the old output layer and copied directly into the corresponding nodes of the new layer, while the weights for the four remaining units remained randomly initialized. The classifier was then trained in two distinct phases:

- **Phase I: Training the new classifier head.** All layers of the base model, including the frozen encoder and the classifier's convolutional and fully-connected blocks, were kept frozen. Only the new softmax output layer was trained on the 10-class dataset, using the same partitioning scheme for the data as the initial 6-class dataset. This first phase rapidly establishes a baseline mapping from the powerful, fixed features to the full set of ten classes, without disturbing the learned weights of the previous layers.
- **Phase II: Fine-tuning the classifier layers.** After Phase I completed, the best-performing model was reloaded from its checkpoint. To refine the model's performance on the more difficult classes, the classifier head was also unfrozen (while maintaining a frozen encoder). The model was then re-compiled with a significantly lower learning rate and trained for additional epochs. This fine-tuning process allows the higher-level decision-making layers to adjust their weights, refining the class boundaries to better accommodate the more subtle distinctions present in the new expanded movement set.

As with the previous training procedures, the same optimization method and callbacks were used during this fine-tuning stage, and the final model's performance was subsequently evaluated on the held-out test set.

## Results

To evaluate the performance of the models, the F1-score metric was used and computed both as a micro-average across all classes, in order to provide an aggregate measure of performance, and on a per-class basis to evaluate the model's efficiency for each individual type of movement. The F1-score is the harmonic mean of precision (P) and recall (R), providing a balanced measure that considers both false positives and false negatives:

$$F_1 = 2 \cdot \frac{P \cdot R}{P + R}. \tag{5}$$

Precision is the ratio of correctly predicted positive observations to the total predicted positive observations:

$$P = \frac{TP}{TP + FP}, \tag{6}$$

where TP (true positives) are the correctly predicted positive instances, while FP (false positives) are the incorrectly predicted positive instance (type I error). Recall, also known as sensitivity or true positive rate, is the ratio of correctly predicted positive observations to all observations in the actual class:

$$R = \frac{TP}{TP + FN}, \tag{7}$$

where FN (false negatives) are the incorrectly predicted negative instances when they were positive (type II error).

To robustly evaluate the performance of the models, a leave-one-subject-out cross-validation (CV) was performed. The presented results are expressed as an average F1-score derived from each of the eight distinct folds. This average F1-score serves as a robust and unbiased indicator of the model's overall performance, reflecting its generalization capability. Additionally, this average is augmented by the inclusion of the standard error (SE) of the mean, which quantitatively characterizes the variability and statistical confidence associated with the reported performance. This comprehensive reporting format results in a thorough understanding of both the central tendency and consistency of the model's performance under rigorous cross-validation.

**Hyperparameter Tuning**

In order to determine the optimal configuration for the feature extractor, a hyperparameter search was conducted on the CSAE architecture. Specifically, we analyzed the impact of two key hyperparameters on the final classification performance: the number of filters in the bottleneck layer and the L1 regularization coefficient $\lambda$. Multiple autoencoders were trained across a grid of these parameters and the features from each trained encoder were then used to train the same classifier architecture. For clarity, only the most insightful results from this search are presented. After identifying the optimal number of filters which consistently yielded the best-performing feature sets, we analyzed the specific impact of $\lambda$ for that configuration. The relationship between the L1 coefficient used in the autoencoder and the final classifier's F1-score is illustrated in Figure 6.

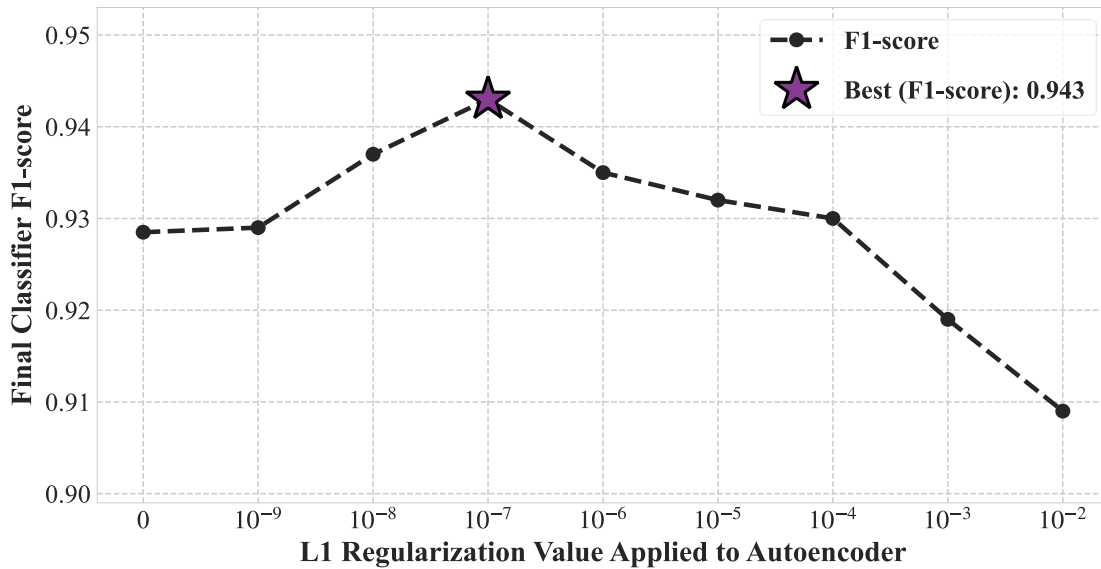

**Figure 6.** Impact of autoencoder sparsity coefficient on final classification performance. The F1-score of the final classifier is shown for various L1 regularization coefficients (λ) applied to the feature-extracting autoencoder. The results demonstrate a clear optimal point for regularization (purple star), with the best performance achieved at $\lambda = 10^{-7}$.

It can be observed that a small amount of regularization $\lambda = 10^{-7}$ yields the peak F1-score, although increasing the regularization strength beyond this point leads to a consistent and significant degradation in performance. This suggests that while a minimal L1 penalty is beneficial, the classification task benefits more from a rich, less-constrained feature representation than from a heavily sparse one. Based on this analysis, the autoencoder configured with $\lambda = 10^{-7}$ was selected as the optimal feature extractor for all subsequent experiments presented in this paper.

**Multi-Subject Classification Performance**

For each of the eight CV folds, a model was trained using data from the remaining seven subjects and then evaluated on the held-out test set (trial 6) from those same subjects. This protocol rigorously measures the model's ability to robustly classify unseen signals from users within the training cohort. The performance metrics reported in Table 1 are the average results across all eight folds. From the presented results we can observe that the model achieved an excellent average F1-score of 94.3% ± 0.3%. The consistency of this high score, indicated by the low standard error calculated across the folds, demonstrates that the proposed feature extraction and classification pipeline can reliably and consistently distinguish between the six movements when a user's general signal patterns are represented in the training data.

This level of performance is highly competitive with established benchmarks for sEMG classification[17,28]. Notably, when compared to our previous research where the feature extraction procedure was done manually by generating well-known statistical time and frequency domain features[29], the proposed deep learning approach achieves comparable accuracy while significantly reducing computational cost and eliminating the need for domain-specific feature engineering.

**Table 1.** Evaluation of multi-subject classification (cross-validation mean F1-score ± standard error)

| Movement | F1-score |
|---|---|
| Hand-close | 96.3% ± 0.5% |
| Thumb | 91.5% ± 0.6% |
| Index finger | 90.2% ± 0.5% |
| Middle finger | 95.0% ± 0.2% |
| Ring finger | 97.8% ± 0.2% |
| Little finger | 95.2% ± 0.5% |
| **Average** | **94.3% ± 0.3%** |

**User-Specific Adaptation**

Next, we evaluated the model's ability to adapt to a new user, where we employed the same leave-one-subject-out cross-validation protocol. In this scheme, the initial model trained on data from seven subjects was tested on the remaining, completely unseen subject to get a pre-tuning score. Subsequently, the model underwent the fine-tuning procedure using only a minimal amount of data (two measurements from each type of movement) from that same unseen subject. This entire process was repeated eight times, with each subject serving as the "new user" once.

The results of the fine-tuning procedure, presented in Table 2, are highly significant. The performance of the original, un-tuned model on the new subject is poor, with an average F1-score of only 35.1% ± 3.1%. The near-zero scores for several movements, such as the flexion of the thumb and index finger, are a clear illustration of the well-known challenge of inter-subject variability in sEMG signals, a fundamental obstacle in developing generalized myoelectric controllers[30]. This demonstrates that a model trained on a group of users cannot be expected to perform reliably on a new individual without some level of adaptation.

**Table 2.** Evaluation of user-specific adaptation (cross-validation mean F1-score $\pm$ standard error)

| Movement | F1-score (original) | F1-score (fine-tuned) |
|---|---|---|
| Hand-close | 42.9% $\pm$ 9.5% | 96.3% $\pm$ 1.4% |
| Thumb | 13.2% $\pm$ 7.7% | 89.3% $\pm$ 2.9% |
| Index finger | 24.2% $\pm$ 11.0% | 89.7% $\pm$ 2.3% |
| Middle finger | 15.9% $\pm$ 4.9% | 91.4% $\pm$ 3.1% |
| Ring finger | 42.6% $\pm$ 10.5% | 95.3% $\pm$ 1.5% |
| Little finger | 30.2% $\pm$ 6.3% | 91.2% $\pm$ 2.6% |
| **Average** | **35.1% $\pm$ 3.1%** | **92.3% $\pm$ 0.9%** |

However, after applying the fine-tuning protocol, the model's performance improves dramatically. The average F1-score surges to 92.3% $\pm$ 0.9%, an increase of over 57%. This significant improvement achieved using only a few short measurements from the new user, coupled with the low standard error across all eight folds, provides strong evidence that the proposed transfer learning approach is not just effective but also consistent and user-independent. It validates the proposed approach by demonstrating that while the foundational features learned by the encoder are robust and generalizable, the final decision-making layers of the classifier can be rapidly and efficiently calibrated to a new user's unique neuromuscular patterns. This capability for rapid personalization is critical for practical prosthetic applications, as it greatly reduces the calibration burden on the end-user.

**Knowledge Transfer for New Movement Integration**

Finally, to evaluate the model's ability to incrementally learn new movements, the 6-class classifier was adapted for the more complex 10-class dataset using the two-phase, weight-transfer protocol. The final performance on the held-out test set for this adapted model, evaluated using the same leave-one-subject-out cross-validation framework, is detailed in Table 3.

The initial training of only the new output layer (Phase I) achieved a promising baseline F1-score of 70.8% $\pm$ 0.9%. However, the subsequent fine-tuning of the classifier head (Phase II) provided a significant and critical improvement, boosting the final average F1-score to 90.0% $\pm$ 0.2%. To validate the necessity of the two-phase approach, we observed that omitting the staged protocol resulted in an average performance drop of ~2%.

This demonstrates that while the base features are highly transferable, allowing the model to refine its higher-level decision boundaries with a low learning rate is essential for accommodating the subtle distinctions in the expanded movement set. The benefit of this fine-tuning process is most pronounced in the classification of the new, more subtle combination movements. For instance, the F1-score for the combined thumb-index finger flexion increased by over 39% (from 53.0% $\pm$ 2.6% to 92.7% $\pm$ 0.5%), while the score for the combined thumb-little finger flexion increased by approximately 29% (from 55.6% $\pm$ 1.7% to 84.9% $\pm$ 0.7%).

**Table 3.** Evaluation of adapted model on expanded 10-class dataset (cross-validation mean F1-score $\pm$ standard error)

| Movement | F1-score (phase I) | F1-score (phase II) |
|---|---|---|
| Hand-close | 86.2% $\pm$ 0.7% | 95.7% $\pm$ 0.4% |
| Thumb | 86.6% $\pm$ 1.3% | 91.2% $\pm$ 0.4% |
| Index finger | 75.4% $\pm$ 1.2% | 87.6% $\pm$ 0.5% |
| Middle finger | 85.3% $\pm$ 1.7% | 94.1% $\pm$ 0.4% |
| Ring finger | 77.6% $\pm$ 1.3% | 90.1% $\pm$ 0.5% |
| Little finger | 57.7% $\pm$ 1.1% | 81.4% $\pm$ 0.7% |
| Thumb-Index | 53.0% $\pm$ 2.6% | 92.7% $\pm$ 0.5% |
| Thumb-Middle | 63.8% $\pm$ 2.5% | 93.9% $\pm$ 0.4% |
| Thumb-Ring | 62.2% $\pm$ 1.7% | 88.8% $\pm$ 0.6% |
| Thumb-Little | 55.6% $\pm$ 1.7% | 84.9% $\pm$ 0.7% |
| **Average** | **70.8% $\pm$ 0.9%** | **90.0% $\pm$ 0.2%** |

**Comparative Analysis with Benchmark Models**

In order to validate the architectural choices of our proposed pipeline, we conducted a final comparative analysis against three benchmark models. The goal was to quantify the performance benefits gained from (1) the automated feature learning approach compared to traditional methods, (2) using a convolutional architecture that preserves temporal information, and (3) the implementation of sparsity in our autoencoder. The features generated by the convolutional autoencoders were evaluated using the same optimized classifier head, while a Random Forest (RF) classifier was employed for the benchmarks that produce a flat feature vector. The final F1-scores for each method on the 6-class dataset are summarized in Table 4.

**Table 4.** Comparative analysis with benchmark feature extraction methods

| Method | F1-score (CV Mean $\pm$ SE) | Static Memory | Runtime Memory | GFLOPs |
|---|---|---|---|---|
| FCAE + Random Forest | 49.7% $\pm$ 0.3% | 9.1 MB | 48 KB | 0.005 |
| Classical Features + Random Forest | 82.0% $\pm$ 0.4% | N/A | 16 KB | 0.00007 |
| CAE | 92.9% $\pm$ 0.4% | 0.74 MB | 0.93 MB | 0.3 |
| **Proposed CSAE** | **94.3% $\pm$ 0.3%** | **0.74 MB** | **0.93 MB** | **0.3** |

*Comparison with a Classical Approach*

To benchmark our deep learning approach against traditional methods, a pipeline using hand-crafted features was implemented. We extracted a standard set of time-domain features (mean absolute value, variance, zero crossings, etc.) from each signal segment and used these to train a RF classifier. The performance of this classical pipeline was expectedly lower than our proposed method. This result clearly demonstrates the power of automated, data-driven feature learning, which can discover complex patterns in the raw sEMG signal that are not captured by pre-defined statistical features.

*Comparison with a Fully-Connected Autoencoder*

Next, we compared our model against a standard fully-connected autoencoder (FCAE) benchmark. The FCAE processes the input by first flattening the (1000, 2) signal segment into a single (2000,) length vector, thereby destroying all temporal information. As shown in Table 4, the classifier trained on features from our proposed CSAE significantly outperformed the one trained on FCAE features, which yielded an F1-score of only $49.7\% \pm 0.3\%$. This result strongly confirms our initial hypothesis that preserving the temporal structure of the sEMG signal via convolutional layers is critical for extracting discriminative features for this task.

*Comparison with a Non-Sparse CAE*

Finally, to isolate the effect of the L1 regularization, we trained an identical CAE architecture but without the sparsity constraint ($\lambda = 0$). While the classifier trained on features from the non-sparse CAE achieved a high F1-score of $92.9\% \pm 0.4\%$, our proposed sparse CAE still yielded a superior result of $94.3\% \pm 0.3\%$. This indicates that even though the convolutional structure provides the primary performance gain, the minimal sparsity penalty provides a beneficial regularizing effect, guiding the autoencoder to learn a slightly cleaner and more effective feature representation.

*Analysis of Computational and Memory Efficiency*

A final analysis of computational and memory requirements is crucial for determining each model's suitability for deployment in a resource constrained embedded system. The chosen metrics quantify the model's total storage requirement (static memory), peak RAM usage (runtime memory), and computational demand (GFLOPs - Giga Floating-Point Operations Per Second). This analysis reveals a clear trade-off between resource efficiency and classification performance. While the classical feature pipeline is exceptionally fast and lightweight, this efficiency comes at the cost of a significant compromise in accuracy. Similarly, the FCAE's low runtime cost is negated by its poor F1-score of $49.7\% \pm 0.3\%$ and a significantly large 9.1 MB static memory requirement, rendering it impractical for memory-limited hardware. The proposed CSAE and the non-sparse CAE prove to be the most balanced, sharing identical and practical resource requirements. This confirms that the CSAE not only delivers state-of-the-art accuracy but achieves it with resource requirements feasible for modern embedded systems, establishing it as the most effective and efficient architecture overall.

## Discussion

The experimental results demonstrate that the proposed pipeline, which leverages a convolutional sparse autoencoder for feature extraction and a lightweight classifier for movement recognition, is a highly effective and adaptable solution for sEMG-based prosthesis control. The findings of this study can be analyzed from three critical perspectives: the model's ability to generalize and adapt, the justification for its specific architecture, and its standing in comparison to existing methodologies.

**Interpretation of Key Findings**

The initial pooled-subject evaluation established a strong performance baseline, with the model achieving an average F1-score of 94.3% ± 0.3% on unseen trials from users present in the training set. This confirms that the CSAE effectively learns a rich and discriminative feature representation capable of reliably classifying movements when a user's general sEMG patterns are known.

More critically, the user-adaptation experiment quantifies the well-known challenge of inter-subject variability. The stark drop in performance to 35.1% ± 3.1% on an unseen user, followed by a surge to 92.3% ± 0.9% after a minimal fine-tuning protocol, is a central finding of this paper. It proves that while a generalized feature extractor is powerful, a mechanism for rapid personalization is not merely beneficial but essential for practical deployment. This result underscores the strength of our transfer learning approach, where a robust foundational model can be efficiently calibrated to an individual's unique neuromuscular patterns with very limited data.

The adaptability of the model was confirmed in the knowledge transfer experiment, where the classifier was successfully updated to classify ten movements instead of six, achieving a final F1-score of 90.0% ± 0.2% without the need for complete retraining. To further evaluate the effectiveness of the adaptation protocol and assess for potential catastrophic forgetting, we directly compared the F1-scores for the six original movements before and after the model was adapted to the expanded 10-class problem. The results of this comparison are illustrated in Figure 7.

As the figure demonstrates, the adapted model largely retains its high performance on most of the original six classes. While there is only a minor performance degradation in some classes, such as the hand-close gesture and flexion of the thumb or middle finger, the F1-scores noticeably drop for flexions of the ring and little finger. Analysis of the aggregate confusion matrices reveals the precise source of this degradation. The performance drop is driven by hierarchical ambiguity between the single and combined movements. Specifically, 82.1% of misclassified ring finger instances were predicted as thumb-ring gestures, and 59.0% of little finger errors were mislabeled as thumb-little movements. Conversely, 87.2% of thumb-ring errors were misclassified as simple ring flexions. This confirms that the primary challenge in the 10-class expansion is not distinguishing distinct fingers, but rather resolving the subtle myoelectric signature of the thumb's simultaneous activation when superimposed on a dominant finger flexion in a low-density sensor setup.

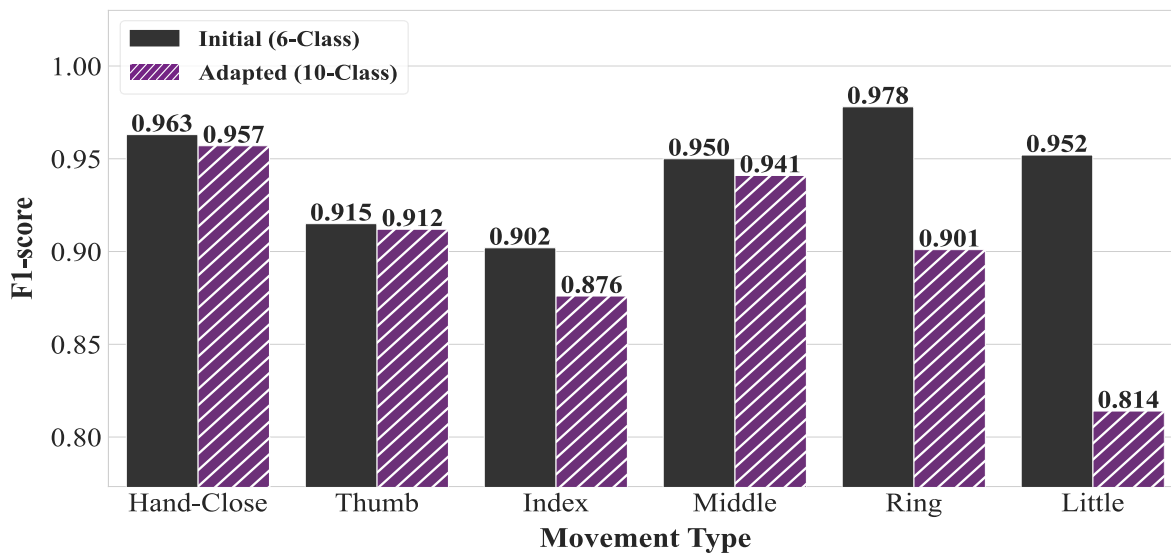

**Figure 7.** F1-score comparison for original movements before and after model adaptation. The black bars represent the performance of the initial 6-class classification model, while the purple (hatched) bars indicate the performance after adapting to a 10-class classification problem. Notably, the adapted model shows comparable or slightly reduced performance for some movements (e.g., hand-close, thumb, middle) but exhibits a more significant decrease in F1-score for others (e.g., index, ring, little) when transitioning from 6 to 10 classes. This highlights the impact of increased class complexity on the classification performance for the initial movement types.

The Phase II fine-tuning procedure, which unfreezes the classifier head (while maintaining a frozen encoder), must adjust the decision boundaries to accommodate the four new, similar movements. This process appears to slightly bias the high-level feature mapping, leading to a performance trade-off on the most challenging and similar base classes. This indicates that while the fine-tuning strategy successfully integrated the knowledge of the new movements, it led to a measurable loss of accuracy for certain previously mastered classes. The results also highlight a key challenge in incremental learning for sEMG data which warrants further investigation. Nevertheless, the findings show that the proposed method is generally suitable for incrementally expanding a prosthesis's functionality in real-world applications, though further efforts to mitigate forgetting in specific problematic classes would enhance its overall robustness.

**Architectural Justification and Comparison to Benchmarks**

The comparative analysis in Table 4 provides clear, quantitative evidence justifying our specific architectural choices. The most significant result is the 44.6% performance gap between our proposed Convolutional Autoencoder and the Fully-Connected Autoencoder. This finding strongly validates our hypothesis that preserving the temporal structure of the sEMG signal is critical; the FCAE, by flattening the input, discards this essential information and fails to learn a useful representation.

Furthermore, our deep learning pipeline significantly outperformed the classical approach using hand-crafted features and a Random Forest classifier. This suggests that the data-driven features learned by the CSAE capture more complex and discriminative patterns than pre-defined statistical measures.

The foundational importance of the convolutional architecture is highlighted by the strong performance of the non-sparse CAE, which achieved a high F1-score of 92.9% $\pm$ 0.4%. However, the fact that our proposed CSAE with a minimal penalty ($\lambda = 10^{-7}$) achieved the highest score of 94.3% $\pm$ 0.3% suggests that a slight sparsity constraint provides a beneficial bias. It likely guides the network to learn a cleaner, more disentangled feature set without discarding the rich, distributed information that, as the L1 tuning curve in Figure 6 showed, is crucial for this task.

**Comparison with Existing Methodologies**

When contextualized within the existing literature, our approach offers a compelling blend of high performance and practical efficiency. While other works have reported high accuracies, they often rely on systems with a larger number of EMG channels or complex preprocessing steps[22,7]. For instance, our model achieves a 94.3% $\pm$ 0.3% F1-score using only two channels, which is a significant improvement over the 85.5% accuracy reported in[22] that required eight EMG channels. Similarly, another recent approach[23] using a Variational Autoencoder (VAE) to disentangle features also required eight channels to achieve its 90.52% cross-subject accuracy. Furthermore, unlike methods that depend on transforming signals into spectrograms or wavelets to achieve high performance[5], our method achieves its results using raw signals, which significantly lowers the potential hardware cost and computational burden. The combination of high accuracy, minimal sensor requirements, and a proven protocol for rapid user adaptation and incremental learning, positions this work as a viable blueprint for low-cost, next-generation prosthetic control systems.

**Feasibility for Real-Time Deployment**

A critical consideration for the translation of this research into a clinical or commercial product is the feasibility of real-time deployment. The computational and memory analysis presented in Table 4 provides strong evidence in this regard.

The proposed CSAE pipeline is highly efficient, requiring only 0.74 MB of static memory and 0.93 MB of runtime memory. This low footprint makes it eminently suitable for on-board, resource-constrained microcontrollers common in commercial prosthetics. In terms of computational load, the model's inference requirement is 0.3 GFLOPS. This efficiency is a direct result of our framework's design. The computationally intensive CSAE encoder is trained offline and then frozen. For real-time inference, the system only needs to execute a single forward pass of this frozen encoder and the very lightweight classifier head. This computational demand is well within the capabilities of modern embedded processors, allowing for classification decisions to be made well within the 250ms window (and supporting the 125ms decision rate).

To assess the computational load, we measured inference latency on a standard consumer CPU (AMD Ryzen 9 5900X). To accurately simulate a resource-constrained embedded environment, the evaluation was restricted to single-core execution. Under these conditions, the model achieved an average latency of 6.4ms per sample, confirming its suitability for low-power microcontrollers. While standard microcontrollers operate at lower clock speeds than workstation CPUs, this low baseline latency, combined with the sampling time of 125ms per data segment, suggests the model is well within the 250-300ms real-time window required for prosthetic control[25]. Furthermore, standard optimization techniques for embedded deployment, such as post-training quantization (int8) and TensorFlow Lite conversion, are known to reduce inference latency and memory footprint by approximately 3–4 times, ensuring that the model remains performant even on resource-constrained hardware such as the ESP32 or Cortex-M series microcontrollers.

This combination of high accuracy, low memory usage, and minimal computational load confirms that the proposed framework is not just a theoretical model but a practical and viable solution for next-generation commercial prosthetic systems.

## Conclusion

This paper has presented a unified, end-to-end framework for myoelectric control that bridges the gap between high-performance deep learning and the strict resource constraints of prosthetic hardware. We have demonstrated that our CSAE-based architecture achieves a multi-subject F1-score of 94.3% using only two sEMG channels. This result challenges the prevailing industry reliance on high-density sensor arrays. Our findings confirm two critical technical hypotheses: (1) preserving temporal structure via 1D convolutions is essential for interpreting raw neuromuscular signals and (2) L1 sparsity constraints serve as a powerful regularizer that forces the model to learn disentangled features that generalize across subjects. Furthermore, our dual-adaptive strategy resolves the limitations of static control architectures. By enabling few-shot adaptation to new users with over 92% accuracy and incremental class expansion with 90% accuracy, the system minimizes the exhaustive calibration typically required of patients. These capabilities highlight the potential of the framework for real-time deployment in resource-constrained environments such as portable embedded systems and microcontrollers.

The promising results of this study open several avenues for future research. To build upon the foundational work with able-bodied subjects, the next logical step is to validate this framework on a larger and more diverse group, including the target population of amputees, to establish its clinical relevance. Furthermore, future work should focus on enhancing the system's robustness by evaluating its performance against real-world challenges, such as muscle fatigue, environmental noise, and electrode shift, a factor known to significantly degrade classifier performance. Finally, expanding the system's capabilities from classifying discrete, static gestures to enabling continuous, proportional control would represent a significant step towards creating more natural and intuitive neuroprosthetic devices, further bridging the gap between artificial limbs and their biological counterparts.

## Data Availability

The dataset analyzed during the current study is a publicly available dataset originally described by Khushaba et al. The data is available from the original authors or can be requested from the corresponding author upon reasonable request.